\documentclass{article}
\usepackage[utf8]{inputenc}
\usepackage[margin=1.0in]{geometry}
\usepackage{amsmath,amssymb}
\usepackage{placeins}
\usepackage{cite}

\usepackage[numbers]{natbib}
\usepackage{graphicx}

\providecommand{\keywords}[1]{\textbf{\textit{Keywords}} #1}

\begin{document}

\title{Scalable Balanced Training of Conditional Generative Adversarial Neural Networks on Image Data}
\author{Massimiliano Lupo Pasini$^1$, Vittorio Gabbi$^2$, Junqi Yin$^3$, Simona Perotto$^4$, Nouamane Laanait$^5$}
\date{}

\maketitle

\noindent $^1$ Oak Ridge National Laboratory, Computational Sciences and Engineering Division, Oak Ridge, TN 37831, USA, email: lupopasinim@ornl.gov\\
$^2$ Politecnico di Milano, Department of Automation and Control Engineering, Milan, MI, 20133, Italy\\
$^3$ Oak Ridge National Laboratory, National Center for Computational Sciences, Oak Ridge, TN, 37831, USA\\
$^4$ Politecnico di Milano, Department of Mathematics, Milan, MI, 20133, Italy\\
$^5$ Anthem, Inc., Atlanta, GA, 30326, USA\

\begin{abstract}
We propose a distributed approach to train deep convolutional generative adversarial neural network (DC-CGANs) models. Our method reduces the imbalance between generator and discriminator by partitioning the training data according to data labels, and enhances scalability by performing a parallel training where multiple generators are concurrently trained, each one of them focusing on a single data label. Performance is assessed in terms of inception score and image quality on MNIST, CIFAR10, CIFAR100, and ImageNet1k datasets, showing a significant improvement in comparison to state-of-the-art techniques to training DC-CGANs.   
Weak scaling is attained on all the four datasets using up to 1,000 processes and 2,000 NVIDIA V100 GPUs on the OLCF supercomputer Summit. 
\end{abstract}
\vspace{0.5cm}

\noindent \keywords{Generative Adversarial Neural Networks \and Deep Learning \and Supercomputing \and Computer Vision}

\vspace{0.5cm}

{\footnotesize \noindent This manuscript has been authored in part by UT-Battelle, LLC, under contract DE-AC05-00OR22725 with the US Department of Energy (DOE). The US government retains and the publisher, by accepting the article for publication, acknowledges that the US government retains a nonexclusive, paid-up, irrevocable, worldwide license to publish or reproduce the published form of this manuscript, or allow others to do so, for US government purposes. DOE will provide public access to these results of federally sponsored research in accordance with the DOE Public Access Plan (http://energy.gov/downloads/doe-public-access-plan).}

\section{Introduction}
\begin{figure}
\centering
\includegraphics[width=\textwidth]{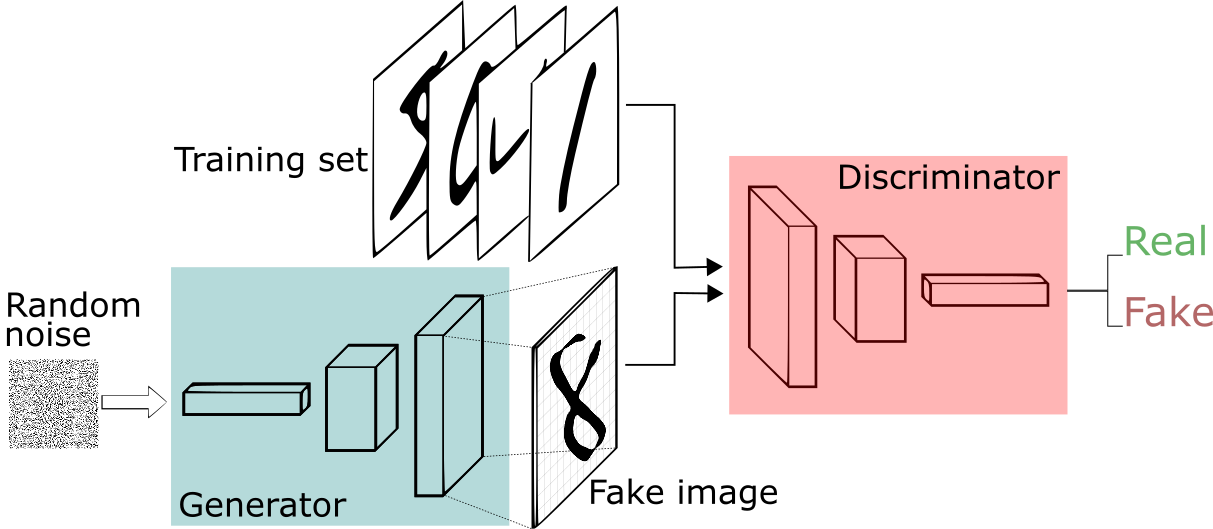}
\caption{The GAN framework pits two adversaries against each other in a game.
Each player is represented by a differentiable function controlled by a set of parameters. Typically these functions are implemented as deep neural networks. Training examples are randomly sampled from the training set and used as input for the first player, the discriminator. The goal of the discriminator is to output the probability that its input is real rather than fake, under the assumption that half of the inputs it is ever shown are real and half are fake. Image from https://sthalles.github.io/intro-to-gans/}
\label{gans_picture}
\end{figure}

Generative adversarial neural networks (GANs) \cite{goodfellow_generative_2014} \cite{radford_unsupervised_2016} \cite{salimans_improved_2016} \cite{bertsekas_multiagent_2019} are deep learning (DL) models whereby a dataset is used by an agent, called the generator, to sample white noise from a latent space and simulate a data distribution to create new (fake) data that resemble the original data it has been trained on. Another agent, called the discriminator, has to correctly discern between the original data (provided by the external environment for training) and the fake data (produced by the generator). The generator prevails over the discriminator if the latter does not succeed in distinguishing anymore the original from the fake. The discriminator prevails over the generator if the fake data created by the generator is categorized as fake, and the original data is still categorized as original. An illustration that describes a GANs model is shown in Figure \ref{gans_picture}.
Originally, GANs have been used on image data to improve the generalizability of DL models for object recognition. Image data are represented as pixels in a Cartesian structure. For each pixel, a set of values called channels are assigned to describe the local graphic properties. The channels per pixel are only one for black and white images, and colored images have three channels (red, green and blue). In general, the graphic variability between classes is more pronounced than the graphic variability within a specific class, because objects of the same type generally resemble more than objects of different nature. The benchmark datasets we consider are characterized by labels that clearly separate images according to their category, and the category is related to the type of object represented in the image. 

The training of GANs is driven by the values of the cost functions associated with the agents \cite{goodfellow_nips_2017}. The cost functions used to evaluate the performance of the discriminator and the generator are related to the number of false positives (original images identified by the discriminator as fake) and false negatives (fake images identified by the discriminator as original). 
The task of the discriminator is relatively simple in that it only has to assign a Boolean value to an image, according to whether the image is predicted as original or fake. On the contrary, the generator needs to map white noise sampled from the latent space into newly created images, and the images created must reproduce relevant features that belong to each data category represented in the training data. 

This imbalance between the difficulty of the computational tasks of discriminator and generator is natural in GANs and cost functions currently used to measure the performance of the generator do not retain information about the disparity in computational tasks between discriminator and generator. 
As a result, the precision attained by the discriminator in performing its tasks (saying if an image is fake or original) is always higher than the precision with which the generator performs its own (create a whole set of fake images from white noise).
Recent game theoretic results show that the unbalanced training of GANs can cause the generator to cycle \cite{mertikopoulos_2017} or converge to a (potentially bad) local optimum \cite{NIPS2017_165a59f7}, which causes the generator to get stuck reproducing only one specific data point (this phenomenon is known in the DL literature as mode collapse). It is thus important to balance the training of GANs models in order to improve the performance of the generator and obtain fake images with similar features to the ones contained in the original data, but this task is challenging \cite{dauphin_equilibrated_2015} \cite{ruder_overview_2017} \cite{duchi_adaptive_nodate} \cite{ward_adagrad_2018} \cite{pmlr-v119-zhao20d}.

Some recent approaches have tackled the imbalance of the two agents by changing the numerical optimization used to train the GANs model. An example is the Competitive Gradient Descent method (CGD) \cite{schafer_competitive_2020}, which recasts the GANs training as a zero-sum game, whereby the discriminator and the generator compete against each other, and the goal is to identify an equilibrium between the agents. However, the zero-sum formulation does not reflect well the interaction between generator and discriminator during the GANs training, since the loss of one agent does not directly translate into the gain of the other agent, as it is indeed assumed in a zero-sum game.   
Other recent approaches have tackled the challenge of imbalance between discriminator and generator by improving the complexity of the GANs model \cite{pascanu_how_2013} \cite{szegedy_going_2014} \cite{sreekumaran_multi-agent_2015} \cite{pmlr-v70-odena17a} \cite{Wang2017GenerativeAN} \cite{karras_progressive_2018} \cite{pmlr-v97-zhang19d} \cite{brock2018large} \cite{zhao2020differentiable}. Among these approaches, Conditional GANs (CGANs) \cite{mirza_conditional_2014} \cite{miyato2018cgans} \cite{DBLP:journals/corr/abs-1912-04216} proceed by expanding the latent space used as input for the generator by adding information about the data categories. The role of CGAN models is to reconstruct a joint data distribution defined on the expanded latent space that combines the image data with the corresponding labels. The inclusion of data labels as an additional latent space variable facilitates the generator in discerning relevant features that make an image more likely to belong to a data category than to another. However, datasets with a large number of categories still pose a non-trivial challenge that prevents the generator from attaining a good performance in creating new fake images still due to a large variability between classes. 

In addition, all the existing approaches to train GANs are characterized by a limited parallelizability, in that existing parallel techniques for GANs are based on data parallelization that distributes large-scale data to multiple replicas of the same model via ensemble learning, and do not further enhance the scalability of GANs by attempting any model parallelization \cite{ruder_overview_2017}. Therefore, state-of-the-art GANs approaches do not fully leverage high-performance computing (HPC) facilities to attain a better performance. 

{\it We propose a novel distributed approach to train CGANs through a non-zero-sum game formulation that uses data categories to address the performance disparity between discriminator and generator and improve the scalability of the CGANs training via model parallelization}. We use the labels to split the data and process each class independently, using a generator for each class. Our distributed approach relies on a factorization of the data distribution where each factor is associated with a single data category. The factorization of the probability distribution makes our approach differ from standard CGANs, where the joint data probability is never decomposed into simpler factors and a single generator is still assigned with the task of creating new images that span all the data categories. The data splitting performed according to the labels removes the variability between classes and thus corrects the imbalance of standard GANs training. 
Because of the independence of each generator from the others, the generators can be trained concurrently and this enhances scalability. 


\section{Distributed conditional generative adversarial neural networks}
\begin{figure}
\centering
\includegraphics[width=\textwidth]{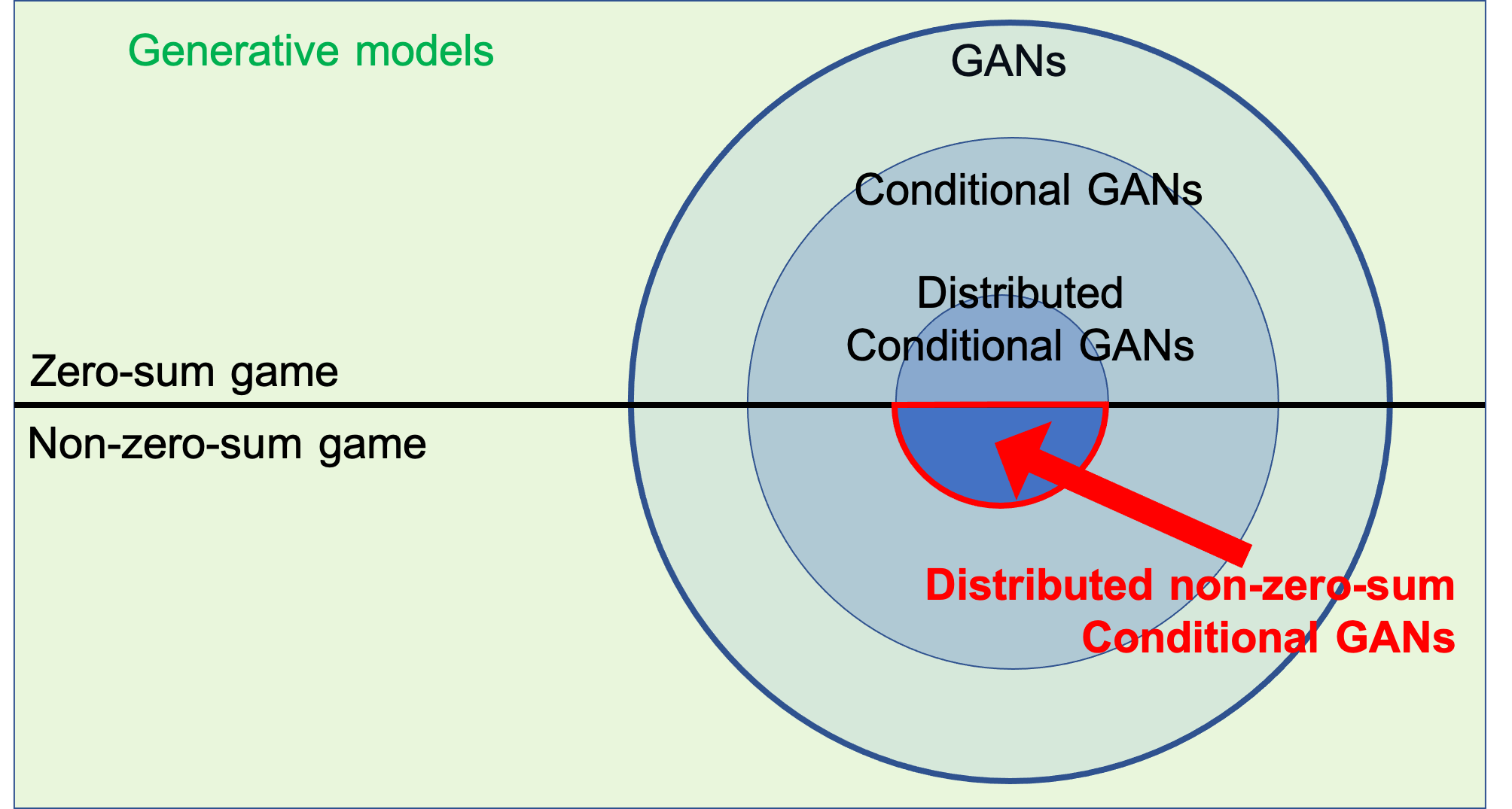}
\caption{Scheme representing the relationship between the generative models currently developed in the deep learning community.}
\label{gans_scheme}
\end{figure}

Our approach aims at implementing a distributed version of CGANs as a non-zero-sum game. Figure \ref{gans_scheme} describes how our approach relates to the other generative models presented in the literature. 
We first define the following input and output spaces, each with an associated probability distribution:
\begin{itemize}
    \item $Z$ is a noise space used to seed the generative model. $Z = \mathbb{R}^{d_Z}$ , where $d_Z$ is a hyperparameter. Values $\mathbf{z} \in Z$ are sampled from a noise distribution $p_\mathbf{z}(\mathbf{z})$. In our experiments $p_\mathbf{z}$ is a white noise model.
    \item $Y$ is an embedding space used to condition the generative model on additional external information, drawn from the training data. $Y = \mathbb{R}^{d_Y}$ , where $d_Y$ is a hyperparameter. Using conditional information provided in the training data, we define a density model $p_\mathbf{y}(\mathbf{y})$.
\item $X$ is the data space which represents an image output
from the generator or input to the discriminator. Values
are normalized pixel values: $X = [0,1]^W \times
C$, where $W$ represents the resolution of the input
images, and $C$ is the set of distinct color channels in
the input images. Using the images in the training data
and their associated conditional data, we can define
a density model $p_{\text{data}}(\mathbf{x})$ of face images. This is
exactly the density model we wish to replicate with the
overall model in this paper.
\end{itemize}
We now define two functions:
\begin{itemize}
\item $G: Z \times Y \rightarrow X$ is the conditional generative model (or generator), which accepts noise data $\mathbf{z}\in Z$ and produces an image $\mathbf{x}\in X$ conditional to the external information $\mathbf{y}\in Y$.
\item $D:X \rightarrow [0, 1]$ is the discriminative model (or
discriminator), which accepts an image $\mathbf{x}$ and condition
$\mathbf{y}$ and predicts the probability under condition $\mathbf{y}$ that $\mathbf{x}$
came from the empirical data distribution rather than
from the generative model.
\end{itemize}
The goal of CGANs is to provide a model that estimates the probability distribution $p_\text{model}(\mathbf{x},\boldsymbol{\theta},\mathbf{y})$, parameterized by parameters $\boldsymbol{\theta}$ that describes the DL model. We then refer to the likelihood as the probability that the model assigns to the training data: $\Pi_{i=1}^m p_\text{model}(\mathbf{x}_i,\boldsymbol{\theta},\mathbf{y})$, for a dataset containing $m$ training samples $\mathbf{x}_i$. Among the different types of generative models, GANs is a type of model that works via the principle of \textit{maximum likelihood}. 
The principle of maximum likelihood aims at choosing the parameters $\boldsymbol{\theta}$ for the DL model that maximize the likelihood of the training data
\begin{equation}
        \boldsymbol{\theta}^* = \underset{\boldsymbol{\theta}}{\operatorname{argmax}}(p_\text{model}(\mathbf{x},\boldsymbol{\theta},\mathbf{y}) ) 
\end{equation}
Using condition information provided in the training data, we define a density model $p_\mathbf{y}(\mathbf{y})$. CGANs use the Bayes theorem to combine the conditional probability $p_\text{model}(\mathbf{x}| \mathbf{y})$ and the density model $p_\mathbf{y}(\mathbf{y})$ to yield the joint model probability $p_\text{model}(\mathbf{x},\mathbf{y})$:
\begin{equation}
    p_\text{model}(\mathbf{x},\boldsymbol{\theta},\mathbf{y}) = p_\text{model}(\mathbf{x},\boldsymbol{\theta}|\mathbf{y})p_\mathbf{y}(\mathbf{y}).
    \label{distributed_conditional}
\end{equation}
While \eqref{distributed_conditional} cannot be expressed in closed analytical form, CGANs can be trained without needing to explicitly define a density function, because this type of generative model offers a way to train the model while interacting only indirectly with $p_\text{model}(\mathbf{x},\boldsymbol{\theta},\mathbf{y})$, usually by sampling from it. 

The two players in the game are represented by two functions. The discriminator is defined by a function $D$ that takes $\mathbf{x}$ as input and uses $\boldsymbol{\theta}^{(D)}$ as parameters. The generator is defined by a function $G$ that takes $\mathbf{z}$ as input and uses $\boldsymbol{\theta}^{(G)}$ as parameters.
Both players have cost functions that are defined in terms of both players’ parameters. 
A consensus has been reached in the literature about which cost functions fully describe the performance of the discriminator due to the simplicity of the discriminator's task \cite{goodfellow_generative_2014} \cite{Lucic2018}, whereas the complexity of the computational task of the generator still keeps different options open as to the cost function that better describes the generator's performance.  

The cost used for the discriminator is
\begin{equation}
    J^{(D)}(\boldsymbol{\theta}^{(D)}, \boldsymbol{\theta}^{(G)}) = - \frac{1}{2} \mathbb{E}_{\mathbf{x}\sim p_{\text{data}}}\log D(\mathbf{x}) - \frac{1}{2} \mathbb{E}_{\mathbf{z}} \log (1-D(G(\mathbf{z})))
    \label{cost_discriminator}
\end{equation}
The cost function we choose for the generator is
\begin{equation}
    J^{(G)}(\boldsymbol{\theta}^{(D)}, \boldsymbol{\theta}^{(G)})  = \frac{1}{2}\mathbb{E}_{\mathbf{z}}\log D(G(\mathbf{z})).
    \label{cost_generator}
\end{equation}
This cost function $J^{(G)}(\boldsymbol{\theta}^{(D)}, \boldsymbol{\theta}^{(G)})$ quantitatively describes the ability of the generator in tricking the discriminator so that the discriminator confuses fake images as real. This choice of $J^{(G)}(\boldsymbol{\theta}^{(D)}, \boldsymbol{\theta}^{(G)})$ better reflects the goal of the generator as an individual agent, with respect to other alternatives that force a strong dependence of $J^{(G)}(\boldsymbol{\theta}^{(D)}, \boldsymbol{\theta}^{(G)})$ on $J^{(D)}(\boldsymbol{\theta}^{(D)}, \boldsymbol{\theta}^{(G)})$, such as, for instance, in the zero-sum game formulation where $J^{(G)}(\boldsymbol{\theta}^{(D)}, \boldsymbol{\theta}^{(G)})=-J^{(D)}(\boldsymbol{\theta}^{(D)}, \boldsymbol{\theta}^{(G)})$. The definition in \eqref{cost_generator} ignores the false positive, because the original images are not a product of the generator (only fake images are). 

The discriminator wishes to minimize $J^{(D)}(\boldsymbol{\theta}^{(D)}, \boldsymbol{\theta}^{(G)})$
and must do so while controlling only $\boldsymbol{\theta}^{(D)}$. The generator wishes to maximize $J^{(G)}(\boldsymbol{\theta}^{(D)}, \boldsymbol{\theta}^{(G)})$ and must do so while controlling only $\boldsymbol{\theta}^{(G)}$.  The solution of this mini-max game is a Nash equilibrium. Here, we use the terminology of local differential Nash equilibria \cite{6736623}. In this context, a Nash equilibrium is a tuple $(\boldsymbol{\theta}^{(D)}, \boldsymbol{\theta}^{(G)})$ that is a local minimum of $J^{(D)}$ with respect to $\boldsymbol{\theta}^{(D)}$ and a local maximum of
$J^{(G)}$ with respect to $\boldsymbol{\theta}^{(G)}$.

The training process consists of a numerical optimization scheme that iteratively updates $\boldsymbol{\theta}^{(D)}$ and $\boldsymbol{\theta}^{(G)}$ \cite{ruder_overview_2017}.
On each step, two minibatches are sampled: a minibatch of $\mathbf{x}$ values from the dataset and a minibatch of $\mathbf{z}$ values drawn from the model’s prior over latent variables. The standard choice of a numerical optimization algorithm used to update $\boldsymbol{\theta}^{(D)}$ and $\boldsymbol{\theta}^{(G)}$ is a gradient-based optimization algorithm called Adam \cite{kingma_adam_2017}. 

\begin{figure}
\center
\includegraphics[width=0.5\textwidth]{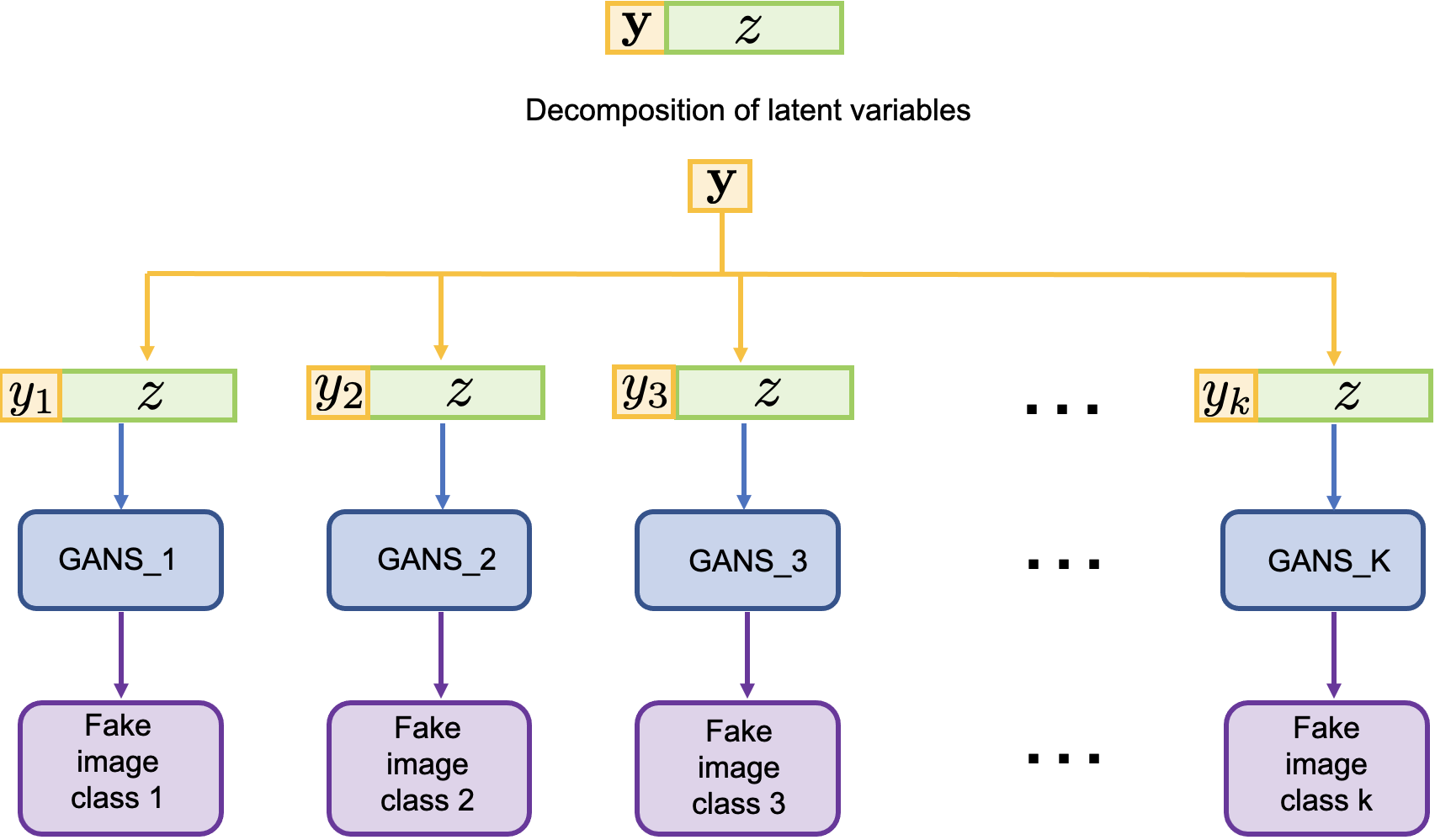}
\caption{Illustration of distributed CGANs.}
\label{distributed_gans_picture}
\end{figure}

Our distributed approach to train CGANs relies on the the equality 
\begin{equation}
    p_\text{model}(\mathbf{x},\boldsymbol{\theta}) = \sum_{k=1}^K p_\text{model}(\mathbf{x},\boldsymbol{\theta}\lvert \mathbf{y}_k)p_\mathbf{y}(\mathbf{y}_k)
\end{equation}
to distribute the computation of each term $p_\text{model}(\mathbf{x},\mathbf{y}_k)$ by training $K$ distributed CGANs, each one per class, and then we combine the results at the end of each training to yield $p_\text{model}(\mathbf{x})$. The numerical examples presented in this paper are characterized by a one-to-one mapping between $\mathbf{y}_k$ and the labels in the image dataset. The advantage of our approach consists in the fact that all the $K$ distributed CGANs can be trained concurrently and independently of each other, thus exposing the model to a higher level of parallelism. If the complexity representation of the objects in each category is comparable, the training time for each distributed CGANs model is approximately the same, which in turn translates into promising performance in terms of weak scalability (the time-to-solution is constant for an increasing number of processors used to solve problems of increasing size so that the computational workload per processor is unchanged). An illustration that describe the distribution of CGANs is provided in Figure \ref{distributed_gans_picture}.

\section{Numerical results}
In this section we compare the performance of standard deep convolutional GANs \cite{radford_unsupervised_2016} (DC-GANs) with deep convolutional conditional GANs (DC-CGANs) \cite{gauthier_conditional_nodate} and our distributed approach to train DC-CGANs.   
The comparison is performed on a quantitative level by measuring the Inception Score (IS) \cite{barratt_note_2018}. The IS takes a list of images and returns a single floating point number, the score.
The score is a measure of how realistic a GAN's output is. IS is an automatic alternative to having humans grade the quality of images.
The score measures two things simultaneously:
the image variety (e.g. each image is a different breed of dog), and whether each image distinctly looks like a real object. If both things are true, the score will be high. If either or both are false, the score will be low.
A higher score describes better performance for GANs, as it means that the GAN model can generate many different distinct and realistic images.
The lowest score possible is zero. Mathematically the highest possible score is infinity, although in practice there will probably emerge a finite ceiling.

The specifics of the architectures for generator and discriminator  used to build DC-GANs models are described in Tables \ref{GANs_generator} and \ref{GANs_discriminator}. The specifics of the architectures for generator and discriminator  used to build DC-CGANs models are described in Tables \ref{CGANs_generator} and \ref{CGANs_discriminator}. The architecture of generator and discriminator for distributed DC-CGANs are the same as for DC-GANs.
The training is performed using the optimizer Adam and a learning rate of $2\mathrm{e}-4$, and a total number of 1,000 epochs for all the types of GANs we consider on each dataset.

\subsection{Hardware description}
The numerical experiments are performed using Summit \cite{summit}, a supercomputer at the Oak Ridge Leadership Computing Facility (OLCF) at Oak Ridge National Laboratory. 
Summit has a hybrid architecture, and each node contains two IBM POWER9 CPUs and six NVIDIA Volta GPUs all connected together with NVIDIA’s high-speed NVLink. Each node has over half a terabyte of coherent memory (high bandwidth memory + DDR4) addressable by all CPUs and GPUs plus 1.6 TB of non-volatile memory (NVMe) storage that can be used as a burst buffer or as extended memory. To provide a high rate of communication and I/O throughput, the nodes are connected in a non-blocking fat-tree using a dual-rail Mellanox EDR InfiniBand interconnect.

\subsection{Software description}
The numerical experiments are performed using \texttt{Python3.7} with \texttt{PyTorch v1.3.1} package \cite{paszke_pytorch_2019} for autodifferentiation to train the DL models with the use of GPUs, and the \texttt{mpi4py v3.0.2} tool is used for distributed computing.

As for the DC-GANs and the DC-CGANs approach, generator and discriminator are mapped to the same MPI processes. As for the distributed DC-CGANs, there are multiple discriminator-generator pairs, each one associated with a specific data class, and every discriminator-generator pair is mapped to an MPI process. Each MPI process instantiated in the distributed DC-CGANs is linked to two GPUs, one dedicated to training the discriminator and one dedicated to training the generator. Therefore, the total number of GPUs used with distributed DC-CGANs amounts to twice the number of MPI processes instantiated. 

\begin{table}
\centering
\begin{tabular}{|c|c|c|c|c|c|c|}
\hline
 \multicolumn{7}{|c|}{Generator} \\
\hline
Layer              & Input dim & Output dim & Activation & Kernel size & stride & padding\\
\hline
Input & 100 & 8192 & ReLU & / & / & /\\
\hline
\multicolumn{7}{|c|}{Resizing} \\
\hline
\multicolumn{7}{|c|}{Batch normalization( epsilon = 1e-5 , momentum = 0.1)} \\
\hline
\multicolumn{7}{|c|}{Upsample(scalefactor = 2)} \\
\hline
Convolution1            & 128 & 128 & / & 3 & 1 & 1\\
\hline
\multicolumn{7}{|c|}{Batch normalization(epsilon = 0.8 , momentum = 0.1)} \\
\hline
\multicolumn{7}{|c|}{leakyReLU(slope = 0.2, inplace=True)} \\
\hline
\multicolumn{7}{|c|}{Upsample(scalefactor = 2)} \\
\hline
Convolution2            & 128 & 64 & / & 3 & 1 & 1\\
\hline
\multicolumn{7}{|c|}{Batch normalization(epsilon = 0.8 , momentum = 0.1)} \\
\hline
\multicolumn{7}{|c|}{leakyReLU(slope = 0.2, inplace=True)} \\
\hline
Convolution3            & 64 & 1 or 3  & / & 3 & 1 & 1\\
\hline
\multicolumn{7}{|c|}{Tanh} \\
\hline
\end{tabular}
\caption{Architecture of the generator in DC-GANs}
\label{GANs_generator}
\end{table}

\begin{table}
\centering
\begin{tabular}{|c|c|c|c|c|c|c|}
\hline
 \multicolumn{7}{|c|}{Discriminator} \\
\hline
Layer              & Input dim & Output dim & Activation & Kernel size & stride & padding\\
\hline
Convolution1            & 3 & 16 & / & 3 & 2 & 1\\
\hline
\multicolumn{7}{|c|}{leakyReLU(slope = 0.2, inplace=True)} \\
\hline
\multicolumn{7}{|c|}{Dropout(0.25)} \\
\hline
Convolution2            & 16 & 32 & / & 3 & 2 & 1\\
\hline
\multicolumn{7}{|c|}{leakyReLU(slope = 0.2, inplace=True)} \\
\hline
\multicolumn{7}{|c|}{Dropout(0.25)} \\
\hline
\multicolumn{7}{|c|}{Batch normalization(epsilon = 0.8 , momentum = 0.1)} \\
\hline
Convolution3            & 32 & 64 & / & 3 & 2 & 1\\
\hline
\multicolumn{7}{|c|}{leakyReLU(slope = 0.2, inplace=True)} \\
\hline
\multicolumn{7}{|c|}{Dropout(0.25)} \\
\hline
\multicolumn{7}{|c|}{Batch normalization(epsilon = 0.8 , momentum = 0.1)} \\
\hline
Convolution4            & 64 & 128 & / & 3 & 2 & 1\\
\hline
\multicolumn{7}{|c|}{leakyReLU(slope = 0.2, inplace=True)} \\
\hline
\multicolumn{7}{|c|}{Dropout(0.25)} \\
\hline
\multicolumn{7}{|c|}{Batch normalization(epsilon = 0.8 , momentum = 0.1)} \\
\hline
output & 2048 & 1 & Sigmoid & / & / & /\\
\hline
\end{tabular}
\caption{Architecture of the discriminator in DC-GANs}
\label{GANs_discriminator}
\end{table}

\begin{table}
\centering
\begin{tabular}{|c|c|c|c|c|c|c|}
\hline
 \multicolumn{7}{|c|}{Generator} \\
\hline
Layer              & Input dim & Output dim & Activation & Kernel size & stride & padding\\
\hline
Input & 110 & 8192 & ReLU & / & / & /\\
\hline
\multicolumn{7}{|c|}{Resizing} \\
\hline
\multicolumn{7}{|c|}{Batch normalization( epsilon = 1e-5 , momentum = 0.1)} \\
\hline
\multicolumn{7}{|c|}{Upsample(scalefactor = 2)} \\
\hline
Convolution1            & 128 & 128 & / & 3 & 1 & 1\\
\hline
\multicolumn{7}{|c|}{Batch normalization(epsilon = 0.8 , momentum = 0.1)} \\
\hline
\multicolumn{7}{|c|}{leakyReLU(slope = 0.2, inplace=True)} \\
\hline
\multicolumn{7}{|c|}{Upsample(scalefactor = 2)} \\
\hline
Convolution2            & 128 & 64 & / & 3 & 1 & 1\\
\hline
\multicolumn{7}{|c|}{Batch normalization(epsilon = 0.8 , momentum = 0.1)} \\
\hline
\multicolumn{7}{|c|}{leakyReLU(slope = 0.2, inplace=True)} \\
\hline
Convolution3            & 64 & 1 or 3  & / & 3 & 1 & 1\\
\hline
\multicolumn{7}{|c|}{Tanh} \\
\hline
\end{tabular}
\caption{Architecture of the generator in DC-CGANs}
\label{CGANs_generator}
\end{table}

\begin{table}
\centering
\begin{tabular}{|c|c|c|c|c|c|c|}
\hline
 \multicolumn{7}{|c|}{Discriminator} \\
\hline
Layer              & Input dim & Output dim & Activation & Kernel size & stride & padding\\
\hline
Convolution1            & 3 & 16 & / & 3 & 2 & 1\\
\hline
\multicolumn{7}{|c|}{leakyReLU(slope = 0.2, inplace=True)} \\
\hline
\multicolumn{7}{|c|}{Dropout(0.25)} \\
\hline
Convolution2            & 16 & 32 & / & 3 & 2 & 1\\
\hline
\multicolumn{7}{|c|}{leakyReLU(slope = 0.2, inplace=True)} \\
\hline
\multicolumn{7}{|c|}{Dropout(0.25)} \\
\hline
\multicolumn{7}{|c|}{Batch normalization(epsilon = 0.8 , momentum = 0.1)} \\
\hline
Convolution3            & 32 & 64 & / & 3 & 2 & 1\\
\hline
\multicolumn{7}{|c|}{leakyReLU(slope = 0.2, inplace=True)} \\
\hline
\multicolumn{7}{|c|}{Dropout(0.25)} \\
\hline
\multicolumn{7}{|c|}{Batch normalization(epsilon = 0.8 , momentum = 0.1)} \\
\hline
Convolution4            & 64 & 128 & / & 3 & 2 & 1\\
\hline
\multicolumn{7}{|c|}{leakyReLU(slope = 0.2, inplace=True)} \\
\hline
\multicolumn{7}{|c|}{Dropout(0.25)} \\
\hline
\multicolumn{7}{|c|}{Batch normalization(epsilon = 0.8 , momentum = 0.1)} \\
\hline
output & 2048 & 1 & None & / & / & /\\
\hline
\end{tabular}
\caption{Architecture of the discriminator in DC-CGANs}
\label{CGANs_discriminator}
\end{table}

\subsection{MNIST}
The ANOVA test run on the MNIST dataset produces a value for the $F$-statistic equal to $38.78$, which leads to a p-value close to zero. This indicates that there is strong statistical evidence to claim that the variability between images of different classes is larger than the variability between images of the same class. The IS obtained with all the three GANs models is shown in Table \ref{mnist_inception}.
The results show a similar performance for DC-GANs, DC-CGANs and distributed DC-CGANs, with a slight improvement using the latter over the other two. 
Samples of fake images generated by the generator of DC-GANs, DC-CGANs ans distributed DC-CGANs trained on MNIST are shown in Figures \ref{MNIST_Adam_CNN}, \ref{MNIST_Adam_Conditional} and \ref{MNIST_Adam_Distributed} respectively. The performance of DC-CGANs is very similar to the one of DC-GANs, because some of the digits are still written in a wobbly manner, preventing a clear understanding of what is the actual digit represented. This happens in situations where multiple digits resemble very much (the shape of a 3 is very similar to the shape of an 8). The distributed DC-CGANs instead has a more consistent and clearer representation of all the digits. 

\begin{table}
\centering
\begin{tabular}{|c|c|c|ll}
\cline{1-3}
 & mean & standard deviation \\ \cline{1-3}
DC-GANs & 2.54 & 0.54 \\ \cline{1-3}
DC-CGANs & 2.62 & 0.03 \\ \cline{1-3}
Distributed DC-CGANs & 2.71 & 0.04 \\ \cline{1-3}
\end{tabular}
\caption{Inception score associated with fake images generated by DC-GANs, DC-CGANs and distributed DC-CGANs trained on MNIST dataset. The inception score is calculated using 10 splits to collect statistics. }
\label{mnist_inception}
\end{table}

\begin{table}
\centering
\begin{tabular}{|c|c|c|c|c|c|}
      \hline
      \includegraphics[width=10mm]{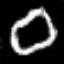} &
      \includegraphics[width=10mm]{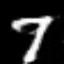} &
      \includegraphics[width=10mm]{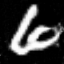} &
      \includegraphics[width=10mm]{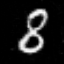} &
      \includegraphics[width=10mm]{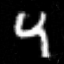} &
      \includegraphics[width=10mm]{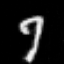} \\
            \hline
      \includegraphics[width=10mm]{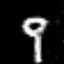} &
      \includegraphics[width=10mm]{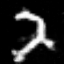} &
\includegraphics[width=10mm]{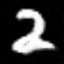} &
      \includegraphics[width=10mm]{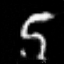} &
      \includegraphics[width=10mm]{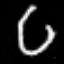} &
      \includegraphics[width=10mm]{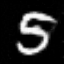} \\
      \hline
      \includegraphics[width=10mm]{MNIST_Adam_sequential/fake_image_MPI_rank_0_Epoch_1107_Batch_300_N_image_7.png} &
      \includegraphics[width=10mm]{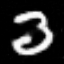} &
      \includegraphics[width=10mm]{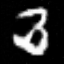} &
      \includegraphics[width=10mm]{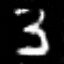} &
      \includegraphics[width=10mm]{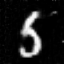} &
      \includegraphics[width=10mm]{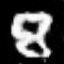} \\
      \hline
      \includegraphics[width=10mm]{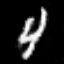} &
      \includegraphics[width=10mm]{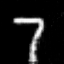} &
      \includegraphics[width=10mm]{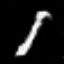} &
      \includegraphics[width=10mm]{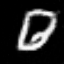} &
      \includegraphics[width=10mm]{MNIST_Adam_sequential/fake_image_MPI_rank_0_Epoch_1108_Batch_400_N_image_7.png} &
      \includegraphics[width=10mm]{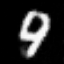} \\
      \hline
      \includegraphics[width=10mm]{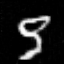} &
      \includegraphics[width=10mm]{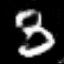} &
      \includegraphics[width=10mm]{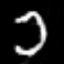} &
      \includegraphics[width=10mm]{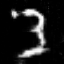} &
      \includegraphics[width=10mm]{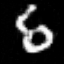} &
      \includegraphics[width=10mm]{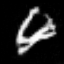} \\
      \hline
      \includegraphics[width=10mm]{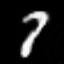} &
      \includegraphics[width=10mm]{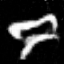} &
      \includegraphics[width=10mm]{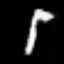} &
      \includegraphics[width=10mm]{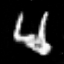} &
      \includegraphics[width=10mm]{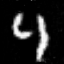} &
      \includegraphics[width=10mm]{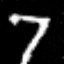} \\
      \hline

\end{tabular}
\caption{Fake images generated by DC-GANs trained on MNIST dataset.}
\label{MNIST_Adam_CNN}
\end{table}

\begin{table}
\centering
\begin{tabular}{|c|c|c|c|c|c|}
      \hline
      \includegraphics[width=10mm]{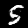} &
      \includegraphics[width=10mm]{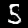} &
      \includegraphics[width=10mm]{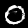} &
      \includegraphics[width=10mm]{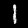} &
      \includegraphics[width=10mm]{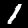} &
      \includegraphics[width=10mm]{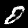} \\
            \hline
      \includegraphics[width=10mm]{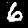} &
      \includegraphics[width=10mm]{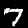} &
\includegraphics[width=10mm]{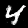} &
      \includegraphics[width=10mm]{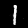} &
      \includegraphics[width=10mm]{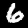} &
      \includegraphics[width=10mm]{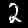} \\
      \hline
      \includegraphics[width=10mm]{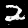} &
      \includegraphics[width=10mm]{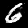} &
      \includegraphics[width=10mm]{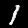} &
      \includegraphics[width=10mm]{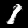} &
      \includegraphics[width=10mm]{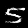} &
      \includegraphics[width=10mm]{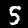} \\
      \hline
      \includegraphics[width=10mm]{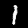} &
      \includegraphics[width=10mm]{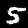} &
      \includegraphics[width=10mm]{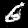} &
      \includegraphics[width=10mm]{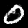} &
      \includegraphics[width=10mm]{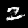} &
      \includegraphics[width=10mm]{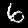} \\
      \hline
      \includegraphics[width=10mm]{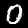} &
      \includegraphics[width=10mm]{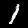} &
      \includegraphics[width=10mm]{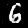} &
      \includegraphics[width=10mm]{MNIST_Adam_conditional/fake_image_MPI_rank_0_Epoch_1516_Batch_0_N_image_10.png} &
      \includegraphics[width=10mm]{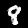} &
      \includegraphics[width=10mm]{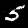} \\
      \hline
      \includegraphics[width=10mm]{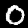} &
      \includegraphics[width=10mm]{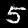} &
      \includegraphics[width=10mm]{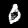} &
      \includegraphics[width=10mm]{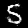} &
      \includegraphics[width=10mm]{MNIST_Adam_conditional/fake_image_MPI_rank_0_Epoch_1516_Batch_500_N_image_1.png} &
      \includegraphics[width=10mm]{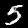} \\
      \hline

\end{tabular}
\caption{Fake images generated by DC-CGANs trained on MNIST dataset.}
\label{MNIST_Adam_Conditional}
\end{table}

\begin{table}
\centering
\begin{tabular}{|c|c|c|c|c|c|}
      \hline
      \includegraphics[width=10mm]{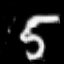} &      
     \includegraphics[width=10mm]{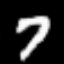} &      
      \includegraphics[width=10mm]{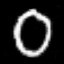} &
      \includegraphics[width=10mm]{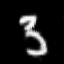} &  
      \includegraphics[width=10mm]{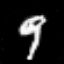} & 
      \includegraphics[width=10mm]{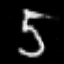} \\      
      \includegraphics[width=10mm]{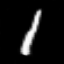} &
      \includegraphics[width=10mm]{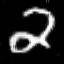} &
      \includegraphics[width=10mm]{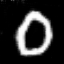} &     
      \includegraphics[width=10mm]{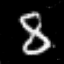} &      
      \includegraphics[width=10mm]{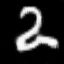} &
      \includegraphics[width=10mm]{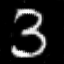} \\
      \includegraphics[width=10mm]{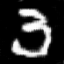} &
     \includegraphics[width=10mm]{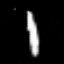} &      
      \includegraphics[width=10mm]{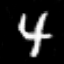} &
      \includegraphics[width=10mm]{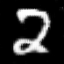} &      
      \includegraphics[width=10mm]{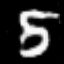} &
      \includegraphics[width=10mm]{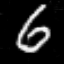} \\
      \includegraphics[width=10mm]{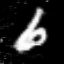} &
      \includegraphics[width=10mm]{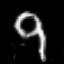} &      
      \includegraphics[width=10mm]{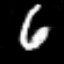} &
      \includegraphics[width=10mm]{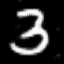} &      
      \includegraphics[width=10mm]{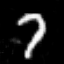} &
      \includegraphics[width=10mm]{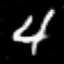} \\     
      \includegraphics[width=10mm]{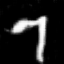} &
      \includegraphics[width=10mm]{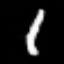} &      
      \includegraphics[width=10mm]{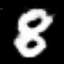} &
      \includegraphics[width=10mm]{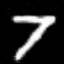} &      
      \includegraphics[width=10mm]{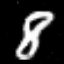} &
     \includegraphics[width=10mm]{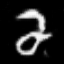} \\      
      \includegraphics[width=10mm]{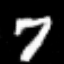} & 
      \includegraphics[width=10mm]{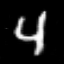} &   
      \includegraphics[width=10mm]{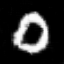} &      
      \includegraphics[width=10mm]{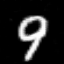} &
      \includegraphics[width=10mm]{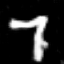} &      
      \includegraphics[width=10mm]{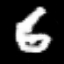}  \\    
      \hline      
      
\end{tabular}
\caption{Fake images generated by distributed DC-CGANs trained on MNIST dataset.}
\label{MNIST_Adam_Distributed}
\end{table}

\subsection{CIFAR10}
The ANOVA test run on the CIFAR10 dataset produces a value for the $F$-statistic equal to $22.12$ for the red channel, $103.20$ for the green channel, and $400.47$ for the blue channel, which lead to p-values close to zero. Also in this case, there is strong statistical evidence to claim that the variability between images of different classes is larger than the variability between images of the same class. 
The IS obtained with all the three GANs models is shown in Table \ref{cifar10_inception}. The distributed approach for DC-CGANs improves the performance of the generator, as it is also shown by a comparison between samples of fake images generated by DC-GANs, DC-CGANs and distributed DC-CGANs models in Figures \ref{CIFAR10_Adam_CNN}, \ref{CIFAR10_Adam_Conditional} and \ref{CIFAR10_Adam_Distributed} respectively. It can be noticed that the images produced with distributed DC-CGANs are more realistic than the ones produced by DC-GANs and standard DC-CGANs, as they represent objects that are easier to recognize as real.

\begin{table}
\centering
\begin{tabular}{|c|c|c|ll}
\cline{1-3}
 & mean & standard deviation \\ \cline{1-3}
DC-GANs & 4.39 & 0.28 \\ \cline{1-3}
DC-CGANs & 5.69 & 0.31 \\ \cline{1-3}
Distributed DC-CGANs & 6.43 & 0.25 \\ \cline{1-3}
\end{tabular}
\caption{Inception score associated with images generated by DC-GANs, DC-CGANs and distributed DC-CGANs trained on CIFAR10 dataset. The inception score is calculated using 10 splits to collect statistics. }
\label{cifar10_inception}
\end{table}

\begin{table}
\centering
\begin{tabular}{|c|c|c|c|c|c|}
      \hline
      \includegraphics[width=10mm]{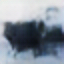} &
      \includegraphics[width=10mm]{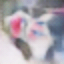} &
      \includegraphics[width=10mm]{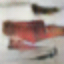} &
      \includegraphics[width=10mm]{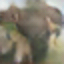} &
      \includegraphics[width=10mm]{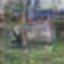} &
      \includegraphics[width=10mm]{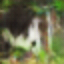} \\
            \hline
      \includegraphics[width=10mm]{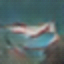} &
      \includegraphics[width=10mm]{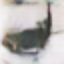} &
\includegraphics[width=10mm]{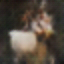} &
      \includegraphics[width=10mm]{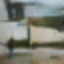} &
      \includegraphics[width=10mm]{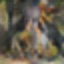} &
      \includegraphics[width=10mm]{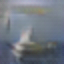} \\
      \hline
      \includegraphics[width=10mm]{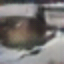} &
      \includegraphics[width=10mm]{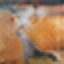} &
      \includegraphics[width=10mm]{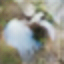} &
      \includegraphics[width=10mm]{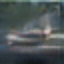} &
      \includegraphics[width=10mm]{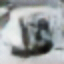} &
      \includegraphics[width=10mm]{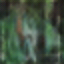} \\
      \hline
      \includegraphics[width=10mm]{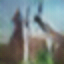} &
      \includegraphics[width=10mm]{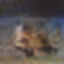} &
      \includegraphics[width=10mm]{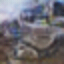} &
      \includegraphics[width=10mm]{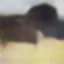} &
      \includegraphics[width=10mm]{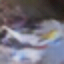} &
      \includegraphics[width=10mm]{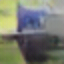} \\
      \hline      \hline
      \includegraphics[width=10mm]{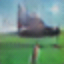} &
      \includegraphics[width=10mm]{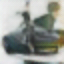} &
      \includegraphics[width=10mm]{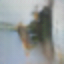} &
      \includegraphics[width=10mm]{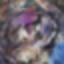} &
      \includegraphics[width=10mm]{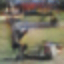} &
      \includegraphics[width=10mm]{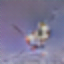} \\
      \hline
      \includegraphics[width=10mm]{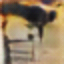} &
      \includegraphics[width=10mm]{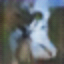} &
      \includegraphics[width=10mm]{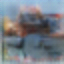} &
      \includegraphics[width=10mm]{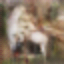} &
      \includegraphics[width=10mm]{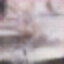} &
      \includegraphics[width=10mm]{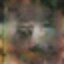} \\
      \hline

\end{tabular}
\caption{Fake images generated by DC-GANs trained on CIFAR10 dataset.}
\label{CIFAR10_Adam_CNN}
\end{table}

\begin{table}
\centering
\begin{tabular}{|c|c|c|c|c|c|}
      \hline
      \includegraphics[width=10mm]{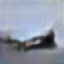} &
      \includegraphics[width=10mm]{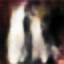} &
      \includegraphics[width=10mm]{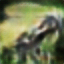} &
      \includegraphics[width=10mm]{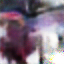} &
      \includegraphics[width=10mm]{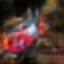} &
      \includegraphics[width=10mm]{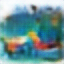} \\
            \hline
      \includegraphics[width=10mm]{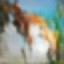} &
      \includegraphics[width=10mm]{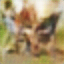} &
\includegraphics[width=10mm]{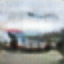} &
      \includegraphics[width=10mm]{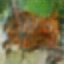} &
      \includegraphics[width=10mm]{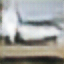} &
      \includegraphics[width=10mm]{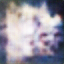} \\
      \hline
      \includegraphics[width=10mm]{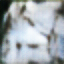} &
      \includegraphics[width=10mm]{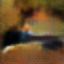} &
      \includegraphics[width=10mm]{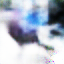} &
      \includegraphics[width=10mm]{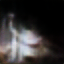} &
      \includegraphics[width=10mm]{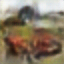} &
      \includegraphics[width=10mm]{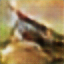} \\
      \hline
      \includegraphics[width=10mm]{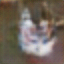} &
      \includegraphics[width=10mm]{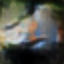} &
      \includegraphics[width=10mm]{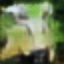} &
      \includegraphics[width=10mm]{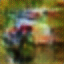} &
      \includegraphics[width=10mm]{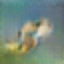} &
      \includegraphics[width=10mm]{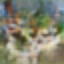} \\
      \hline      \hline
      \includegraphics[width=10mm]{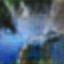} &
      \includegraphics[width=10mm]{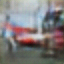} &
      \includegraphics[width=10mm]{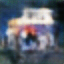} &
      \includegraphics[width=10mm]{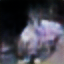} &
      \includegraphics[width=10mm]{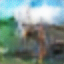} &
      \includegraphics[width=10mm]{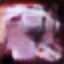} \\
      \hline
      \includegraphics[width=10mm]{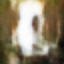} &
      \includegraphics[width=10mm]{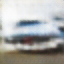} &
      \includegraphics[width=10mm]{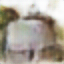} &
      \includegraphics[width=10mm]{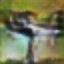} &
      \includegraphics[width=10mm]{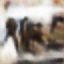} &
      \includegraphics[width=10mm]{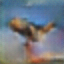} \\
      \hline

\end{tabular}
\caption{Fake images generated by DC-CGANs trained on CIFAR10 dataset.}
\label{CIFAR10_Adam_Conditional}
\end{table}

\begin{table}
\centering
\begin{tabular}{|c|c|c|c|c|c|}
      \hline
      \includegraphics[width=10mm]{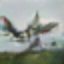} &
      \includegraphics[width=10mm]{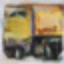} &      
      \includegraphics[width=10mm]{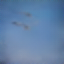} &
      \includegraphics[width=10mm]{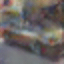} &
      \includegraphics[width=10mm]{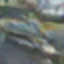} &
      \includegraphics[width=10mm]{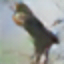} \\
      \includegraphics[width=10mm]{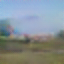} &     
      \includegraphics[width=10mm]{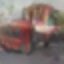} &      
      \includegraphics[width=10mm]{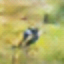} &
      \includegraphics[width=10mm]{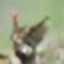} &
      \includegraphics[width=10mm]{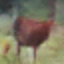} &
      \includegraphics[width=10mm]{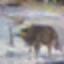} \\
      \includegraphics[width=10mm]{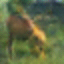} &
      \includegraphics[width=10mm]{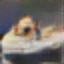} &      
      \includegraphics[width=10mm]{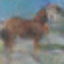} &
      \includegraphics[width=10mm]{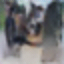} &
      \includegraphics[width=10mm]{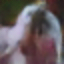} &
      \includegraphics[width=10mm]{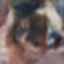} \\
      \includegraphics[width=10mm]{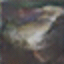} &
      \includegraphics[width=10mm]{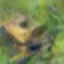} &
      \includegraphics[width=10mm]{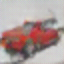} &   
      \includegraphics[width=10mm]{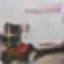} &      
      \includegraphics[width=10mm]{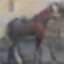} &
      \includegraphics[width=10mm]{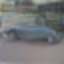} \\      
      \includegraphics[width=10mm]{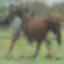} &
      \includegraphics[width=10mm]{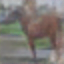} &
      \includegraphics[width=10mm]{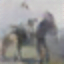} &
      \includegraphics[width=10mm]{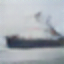} &
      \includegraphics[width=10mm]{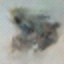} &      
      \includegraphics[width=10mm]{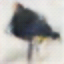} \\      
      \includegraphics[width=10mm]{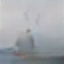} &
      \includegraphics[width=10mm]{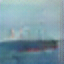} &
      \includegraphics[width=10mm]{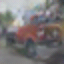} &
      \includegraphics[width=10mm]{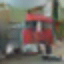} &
      \includegraphics[width=10mm]{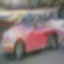} &      
      \includegraphics[width=10mm]{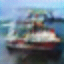} \\
      \hline
\end{tabular}
\caption{Fake images generated by distributed DC-CGANs trained on CIFAR10 dataset.}
\label{CIFAR10_Adam_Distributed}
\end{table}

\subsection{CIFAR100}
The ANOVA test run on the CIFAR100 produces a value for the $F$-statistic equal to $37.43$ for the red channel, $23.74$ for the green channel, and $26.08$ for the blue channel, which lead to p-values close to zero. Also in this case, the variability between images of different classes is larger than the variability between images of the same class. 
The IS obtained with all the three GANs models is shown in Table \ref{cifar100_inception}. The results show an improved performance using DC-CGANs instead of DC-GANs, which is further improved by our distributed approach to train DC-CGANs. 
Samples of fake images generated by the generator of DC-GANs, DC-CGANs and distributed DC-CGANs trained on CIFAR100 are shown in Figures \ref{CIFAR100_Adam_CNN}, \ref{CIFAR100_Adam_Conditional} and \ref{CIFAR100_Adam_Distributed} respectively, and the results still show distributed DC-CGANs generating more realistic pictures that can better be recognized as real objects. 

\begin{table}
\centering
\begin{tabular}{|c|c|c|ll}
\cline{1-3}
 & mean & standard deviation \\ \cline{1-3}
DC-GANs & 4.59 & 0.28 \\ \cline{1-3}
DC-CGANs & 5.62 & 0.23 \\ \cline{1-3}
Distributed DC-CGANs & 6.61 & 0.21 \\ \cline{1-3}
\end{tabular}
\caption{Inception score associated with fake images generated by DC-GANs, DC-CGANs and distributed DC-CGANs trained on CIFAR100 dataset. The inception score is calculated using 10 splits to collect statistics. }
\label{cifar100_inception}
\end{table}

\begin{table}
\centering
\begin{tabular}{|c|c|c|c|c|c|}
      \hline
      \includegraphics[width=10mm]{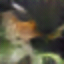} &
      \includegraphics[width=10mm]{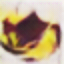} &
      \includegraphics[width=10mm]{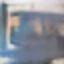} &
      \includegraphics[width=10mm]{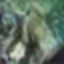} &
      \includegraphics[width=10mm]{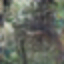} &
      \includegraphics[width=10mm]{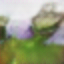} \\
            \hline
      \includegraphics[width=10mm]{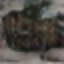} &
      \includegraphics[width=10mm]{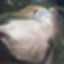} &
\includegraphics[width=10mm]{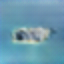} &
      \includegraphics[width=10mm]{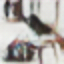} &
      \includegraphics[width=10mm]{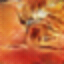} &
      \includegraphics[width=10mm]{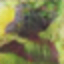} \\
      \hline
      \includegraphics[width=10mm]{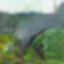} &
      \includegraphics[width=10mm]{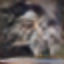} &
      \includegraphics[width=10mm]{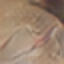} &
      \includegraphics[width=10mm]{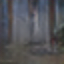} &
      \includegraphics[width=10mm]{CIFAR100_Adam_sequential/fake_image_MPI_rank_0_Epoch_113_Batch_400_N_image_10.png} &
      \includegraphics[width=10mm]{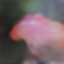} \\
      \hline
      \includegraphics[width=10mm]{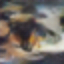} &
      \includegraphics[width=10mm]{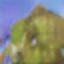} &
      \includegraphics[width=10mm]{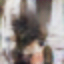} &
      \includegraphics[width=10mm]{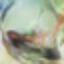} &
      \includegraphics[width=10mm]{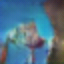} &
      \includegraphics[width=10mm]{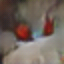} \\
      \hline      \hline
      \includegraphics[width=10mm]{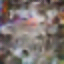} &
      \includegraphics[width=10mm]{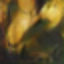} &
      \includegraphics[width=10mm]{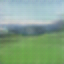} &
      \includegraphics[width=10mm]{CIFAR10_Adam_sequential/fake_image_MPI_rank_0_Epoch_63_Batch_200_N_image_7.png} &
      \includegraphics[width=10mm]{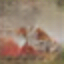} &
      \includegraphics[width=10mm]{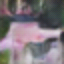} \\
      \hline
      \includegraphics[width=10mm]{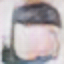} &
      \includegraphics[width=10mm]{CIFAR100_Adam_sequential/fake_image_MPI_rank_0_Epoch_115_Batch_200_N_image_10.png} &
      \includegraphics[width=10mm]{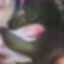} &
      \includegraphics[width=10mm]{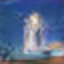} &
      \includegraphics[width=10mm]{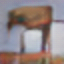} &
      \includegraphics[width=10mm]{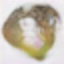} \\
      \hline

\end{tabular}
\caption{Fake images generated by DC-GANs trained on CIFAR100.}
\label{CIFAR100_Adam_CNN}
\end{table}

\begin{table}
\centering
\begin{tabular}{|c|c|c|c|c|c|}
      \hline
      \includegraphics[width=10mm]{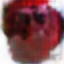} &
      \includegraphics[width=10mm]{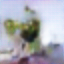} &
      \includegraphics[width=10mm]{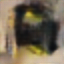} &
      \includegraphics[width=10mm]{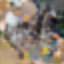} &
      \includegraphics[width=10mm]{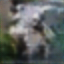} &
      \includegraphics[width=10mm]{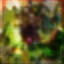} \\
            \hline
      \includegraphics[width=10mm]{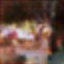} &
      \includegraphics[width=10mm]{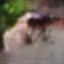} &
     \includegraphics[width=10mm]{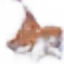} &
      \includegraphics[width=10mm]{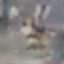} &
      \includegraphics[width=10mm]{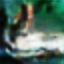} &
      \includegraphics[width=10mm]{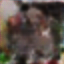} \\
      \hline
      \includegraphics[width=10mm]{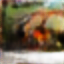} &
      \includegraphics[width=10mm]{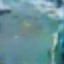} &
      \includegraphics[width=10mm]{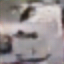} &
      \includegraphics[width=10mm]{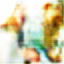} &
      \includegraphics[width=10mm]{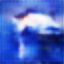} &
      \includegraphics[width=10mm]{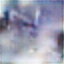} \\
      \hline
      \includegraphics[width=10mm]{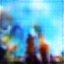} &
      \includegraphics[width=10mm]{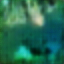} &
      \includegraphics[width=10mm]{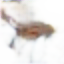} &
      \includegraphics[width=10mm]{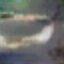} &
      \includegraphics[width=10mm]{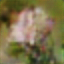} &
      \includegraphics[width=10mm]{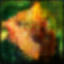} \\
      \hline      \hline
      \includegraphics[width=10mm]{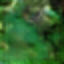} &
      \includegraphics[width=10mm]{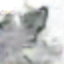} &
      \includegraphics[width=10mm]{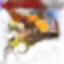} &
      \includegraphics[width=10mm]{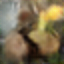} &
      \includegraphics[width=10mm]{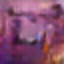} &
      \includegraphics[width=10mm]{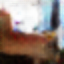} \\
      \hline
      \includegraphics[width=10mm]{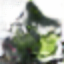} &
      \includegraphics[width=10mm]{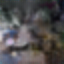} &
      \includegraphics[width=10mm]{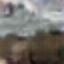} &
      \includegraphics[width=10mm]{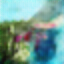} &
      \includegraphics[width=10mm]{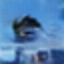} &
      \includegraphics[width=10mm]{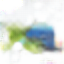} \\
      \hline

\end{tabular}
\caption{Fake images generated by DC-CGANs trained on CIFAR100}
\label{CIFAR100_Adam_Conditional}
\end{table}

\begin{table}
\centering
\begin{tabular}{|c|c|c|c|c|c|}
      \hline
      \includegraphics[width=10mm]{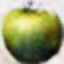} &
      \includegraphics[width=10mm]{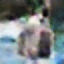} &    
      \includegraphics[width=10mm]{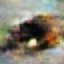} &    
     \includegraphics[width=10mm]{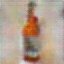} &
      \includegraphics[width=10mm]{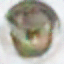} &
      \includegraphics[width=10mm]{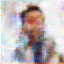} \\  
            \hline
      \includegraphics[width=10mm]{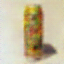} &
      \includegraphics[width=10mm]{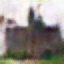} &
      \includegraphics[width=10mm]{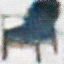} &
      \includegraphics[width=10mm]{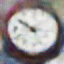} &
      \includegraphics[width=10mm]{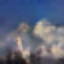} &
      \includegraphics[width=10mm]{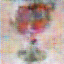} \\
            \hline
      \includegraphics[width=10mm]{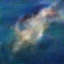} & 
      \includegraphics[width=10mm]{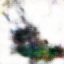} &
      \includegraphics[width=10mm]{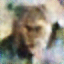} &     
     \includegraphics[width=10mm]{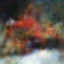} &     
      \includegraphics[width=10mm]{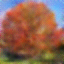} &
      \includegraphics[width=10mm]{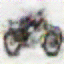} \\
            \hline
      \includegraphics[width=10mm]{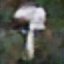} &
      \includegraphics[width=10mm]{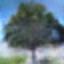} &
      \includegraphics[width=10mm]{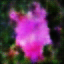} &
      \includegraphics[width=10mm]{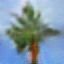} &
      \includegraphics[width=10mm]{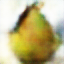} &
      \includegraphics[width=10mm]{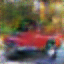} \\
            \hline
      \includegraphics[width=10mm]{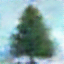} &
      \includegraphics[width=10mm]{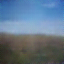} &
      \includegraphics[width=10mm]{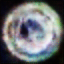} &
      \includegraphics[width=10mm]{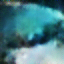} &      
      \includegraphics[width=10mm]{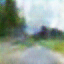} &
     \includegraphics[width=10mm]{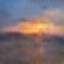} \\ 
     \hline
     \includegraphics[width=10mm]{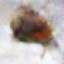} &
      \includegraphics[width=10mm]{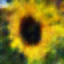} &
      \includegraphics[width=10mm]{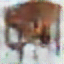} &
      \includegraphics[width=10mm]{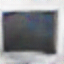} &
      \includegraphics[width=10mm]{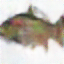} &
     \includegraphics[width=10mm]{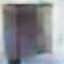} \\     
\end{tabular}
\caption{Fake images generated by distributed DC-CGANs trained on CIFAR100. 
}
\label{CIFAR100_Adam_Distributed}
\end{table}

\subsection{ImageNet1k}
The ANOVA test run on the ImageNet1k produces a value for the $F$-statistic equal to $57.43$ for the red channel, $43.74$ for the green channel, and $56.08$ for the blue channel, which lead to p-values close to zero. The variability between images of different classes is larger than the variability between images of the same class, and this justifies a distributed training for GANs as we propose.  
The IS obtained with all the three GANs models is shown in Table \ref{imagenet_inception}. The results show an improved performance using DC-CGANs instead of DC-GANs, which is further improved by our distributed approach to train DC-CGANs. 
Samples of fake images generated by the generator of DC-GANs, DC-CGANs and distributed DC-CGANs trained on ImageNet1k are shown in Figures \ref{Imagenet_Adam_CNN}, \ref{Imagenet_Adam_conditional} and \ref{Imagenet_Adam_Distributed} respectively. 

\begin{table}
\centering
\begin{tabular}{|c|c|c|ll}
\cline{1-3}
 & mean & standard deviation \\ \cline{1-3}
DC-GANs & 7.24 & 0.78 \\ \cline{1-3}
DC-CGANs & 8.91 & 0.85 \\ \cline{1-3}
Distributed DC-CGANs & 11.61 & 1.12 \\ \cline{1-3}
\end{tabular}
\caption{Inception score associated with fake images generated by DC-GANs, DC-CGANs and distributed DC-CGANs trained on ImageNet1k dataset. The inception score is calculated using 10 splits to collect statistics. }
\label{imagenet_inception}
\end{table}

\begin{table}
\centering
\begin{tabular}{|c|c|c|c|c|c|}
      \hline
      \includegraphics[width=10mm]{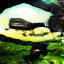} &
      \includegraphics[width=10mm]{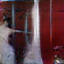} &
      \includegraphics[width=10mm]{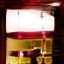} &
      \includegraphics[width=10mm]{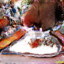} &
      \includegraphics[width=10mm]{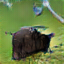} &
      \includegraphics[width=10mm]{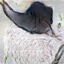} \\
            \hline
      \includegraphics[width=10mm]{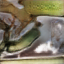} &
      \includegraphics[width=10mm]{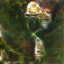} &
     \includegraphics[width=10mm]{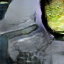} &
      \includegraphics[width=10mm]{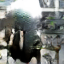} &
      \includegraphics[width=10mm]{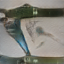} &
      \includegraphics[width=10mm]{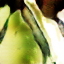} \\
      \hline
      \includegraphics[width=10mm]{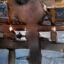} &
      \includegraphics[width=10mm]{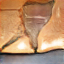} &
      \includegraphics[width=10mm]{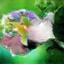} &
      \includegraphics[width=10mm]{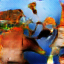} &
      \includegraphics[width=10mm]{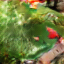} &
      \includegraphics[width=10mm]{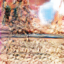} \\
      \hline
      \includegraphics[width=10mm]{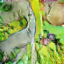} &
      \includegraphics[width=10mm]{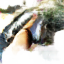} &
      \includegraphics[width=10mm]{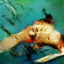} &
      \includegraphics[width=10mm]{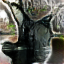} &
      \includegraphics[width=10mm]{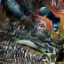} &
      \includegraphics[width=10mm]{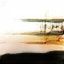} \\
      \hline      \hline
      \includegraphics[width=10mm]{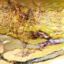} &
      \includegraphics[width=10mm]{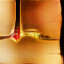} &
      \includegraphics[width=10mm]{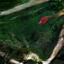} &
      \includegraphics[width=10mm]{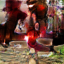} &
      \includegraphics[width=10mm]{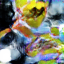} &
      \includegraphics[width=10mm]{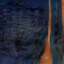} \\
      \hline
      \includegraphics[width=10mm]{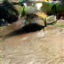} &
      \includegraphics[width=10mm]{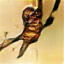} &
      \includegraphics[width=10mm]{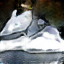} &
      \includegraphics[width=10mm]{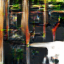} &
      \includegraphics[width=10mm]{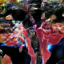} &
      \includegraphics[width=10mm]{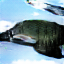} \\
      \hline

\end{tabular}
\caption{Fake images generated by DC-GANs trained on ImageNet1k}
\label{Imagenet_Adam_CNN}
\end{table}

\begin{table}
\centering
\begin{tabular}{|c|c|c|c|c|c|}
      \hline
      \includegraphics[width=10mm]{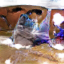} &
      \includegraphics[width=10mm]{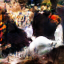} &
      \includegraphics[width=10mm]{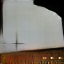} &
      \includegraphics[width=10mm]{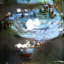} &
      \includegraphics[width=10mm]{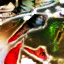} &
      \includegraphics[width=10mm]{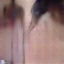} \\
            \hline
      \includegraphics[width=10mm]{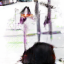} &
      \includegraphics[width=10mm]{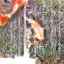} &
     \includegraphics[width=10mm]{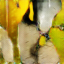} &
      \includegraphics[width=10mm]{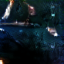} &
      \includegraphics[width=10mm]{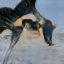} &
      \includegraphics[width=10mm]{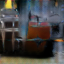} \\
      \hline
      \includegraphics[width=10mm]{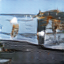} &
      \includegraphics[width=10mm]{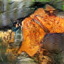} &
      \includegraphics[width=10mm]{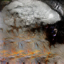} &
      \includegraphics[width=10mm]{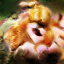} &
      \includegraphics[width=10mm]{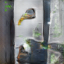} &
      \includegraphics[width=10mm]{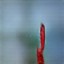} \\
      \hline
      \includegraphics[width=10mm]{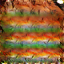} &
      \includegraphics[width=10mm]{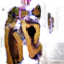} &
      \includegraphics[width=10mm]{ImageNet1k_Adam_conditional/fake_image_MPI_rank_0_Epoch_10_Batch_1700_N_image_10.png} &
      \includegraphics[width=10mm]{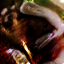} &
      \includegraphics[width=10mm]{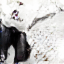} &
      \includegraphics[width=10mm]{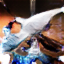} \\
      \hline      \hline
      \includegraphics[width=10mm]{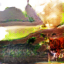} &
      \includegraphics[width=10mm]{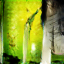} &
      \includegraphics[width=10mm]{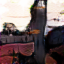} &
      \includegraphics[width=10mm]{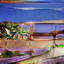} &
      \includegraphics[width=10mm]{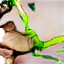} &
      \includegraphics[width=10mm]{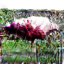} \\
      \hline
      \includegraphics[width=10mm]{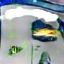} &
      \includegraphics[width=10mm]{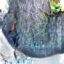} &
      \includegraphics[width=10mm]{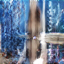} &
      \includegraphics[width=10mm]{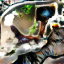} &
      \includegraphics[width=10mm]{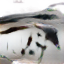} &
      \includegraphics[width=10mm]{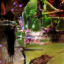} \\
      \hline

\end{tabular}
\caption{Fake images generated by DC-CGANs trained on ImageNet1k}
\label{Imagenet_Adam_conditional}
\end{table}

\begin{table}
\centering
\begin{tabular}{|c|c|c|c|c|c|}
      \hline
      \includegraphics[width=10mm]{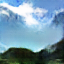} &
      \includegraphics[width=10mm]{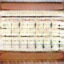} &    
      \includegraphics[width=10mm]{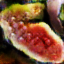} &    
     \includegraphics[width=10mm]{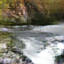} &
      \includegraphics[width=10mm]{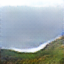} &
      \includegraphics[width=10mm]{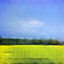} \\  
            \hline
      \includegraphics[width=10mm]{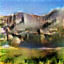} &
      \includegraphics[width=10mm]{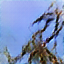} &
      \includegraphics[width=10mm]{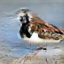} &
      \includegraphics[width=10mm]{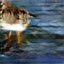} &
      \includegraphics[width=10mm]{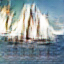} &
      \includegraphics[width=10mm]{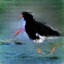} \\
            \hline
      \includegraphics[width=10mm]{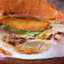} & 
      \includegraphics[width=10mm]{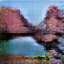} &
      \includegraphics[width=10mm]{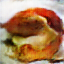} &     
     \includegraphics[width=10mm]{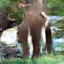} &     
      \includegraphics[width=10mm]{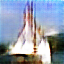} &
      \includegraphics[width=10mm]{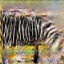} \\
            \hline
      \includegraphics[width=10mm]{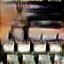} &
      \includegraphics[width=10mm]{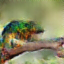} &
      \includegraphics[width=10mm]{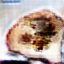} &
      \includegraphics[width=10mm]{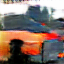} &
      \includegraphics[width=10mm]{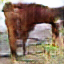} &
      \includegraphics[width=10mm]{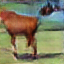} \\
            \hline
      \includegraphics[width=10mm]{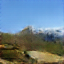} &
      \includegraphics[width=10mm]{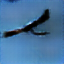} &
      \includegraphics[width=10mm]{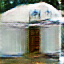} &
      \includegraphics[width=10mm]{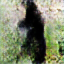} &      
      \includegraphics[width=10mm]{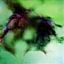} &
     \includegraphics[width=10mm]{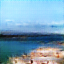} \\ 
     \hline
     \includegraphics[width=10mm]{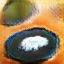} &
      \includegraphics[width=10mm]{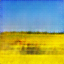} &
      \includegraphics[width=10mm]{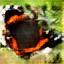} &
      \includegraphics[width=10mm]{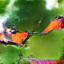} &
      \includegraphics[width=10mm]{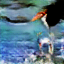} &
     \includegraphics[width=10mm]{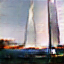} \\   
     \hline
     \includegraphics[width=10mm]{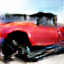} &
      \includegraphics[width=10mm]{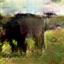} &
      \includegraphics[width=10mm]{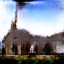} &
      \includegraphics[width=10mm]{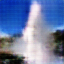} &
      \includegraphics[width=10mm]{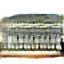} &
     \includegraphics[width=10mm]{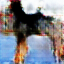} \\         
\end{tabular}
\caption{Fake images generated by distributed DC-CGANs trained on ImageNet1k. 
}
\label{Imagenet_Adam_Distributed}
\end{table}

\subsection{Scaling performance of distributed DC-CGANs}
\begin{figure*}
   \centering
\begin{tabular}{cc}
\includegraphics[width=0.4\textwidth]{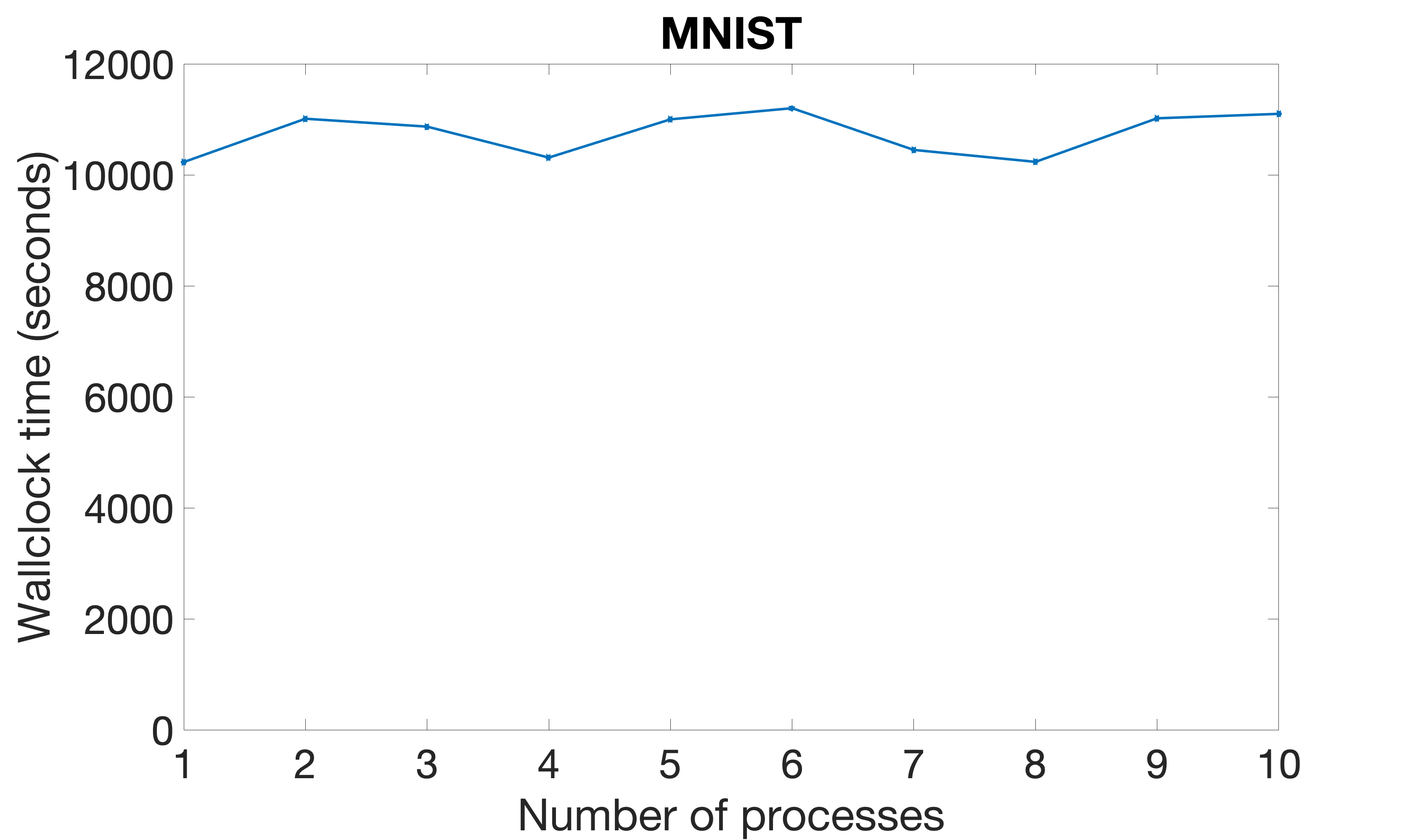}&
\includegraphics[width=0.4\textwidth]{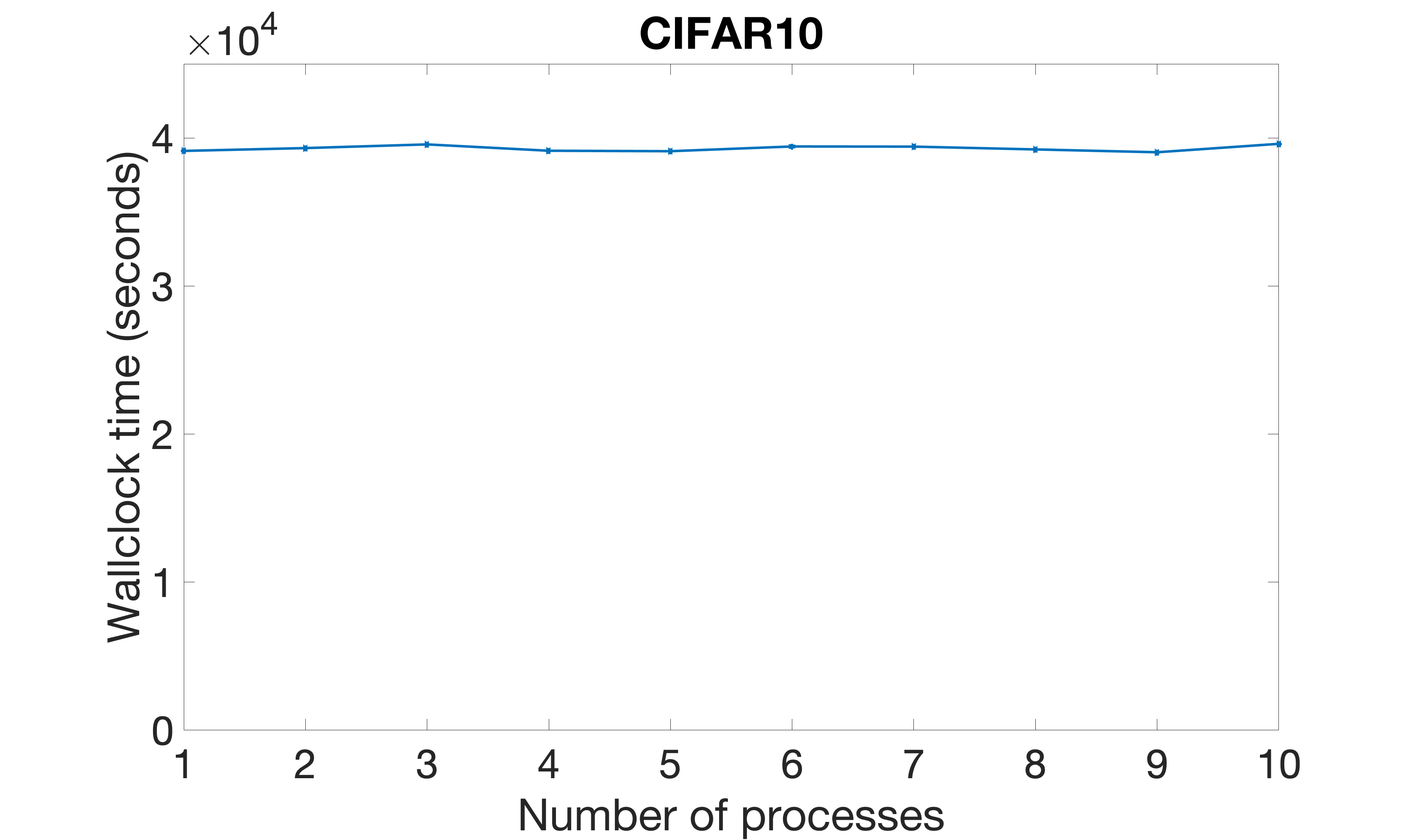}\\
\includegraphics[width=0.4\textwidth]{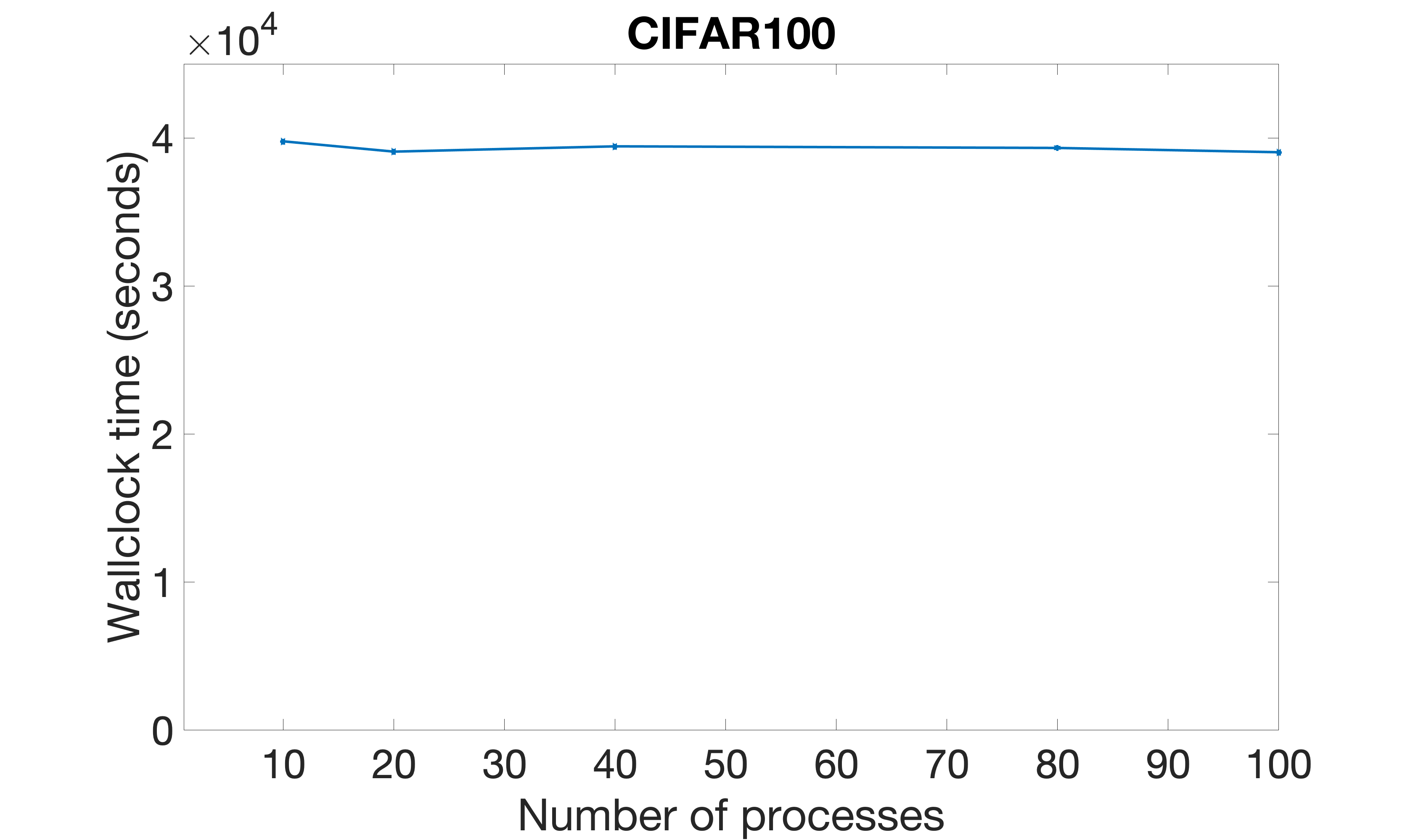}&
\includegraphics[width=0.4\textwidth]{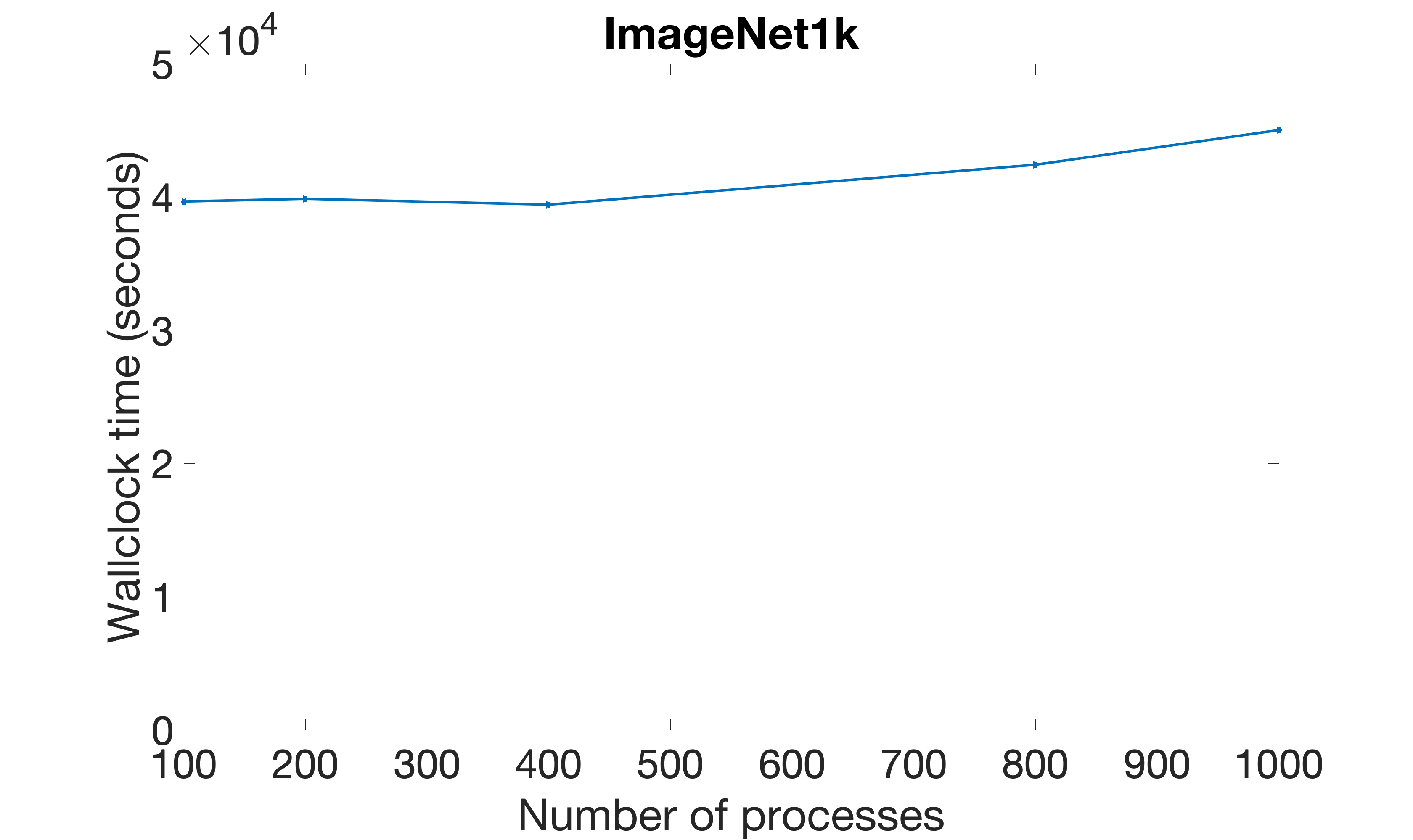}
\\
\end{tabular}

    \caption{Results of weak scaling tests for MNIST, CIFAR10, CIFAR100, and ImageNet1k}
    \label{scalability_pic} 
\end{figure*}

We tested the scalability of our distributed approach to train DC-CGANs by running experiments on OLCF supercomputer Summit. We measured the wall-clock time needed by the distributed DC-CGANs to complete the training as a function of the number of data classes (and thus MPI processes). The parameters for the numerical optimization are the same as discussed before. The training of the model has been performed by distributing the computation through a one-to-one mapping between the \texttt{MPI} process and the data classes. Each MPI process was mapped to two NVIDIA V100 GPUs, so that the neural networks for discriminator and generator for each data class would be trained on separate GPUs. For the MNIST and CIFAR10 datasets, the number of MPI processes spans the range from 1 to 10, whereas the number of MPI processes, for the CIFAR100 dataset was set to 10, 20, 40, 80 and 100, and for the ImageNet1k dataset was set to 100, 200, 400, 800 and 1,000. The results for the scalability tests are shown in Figure \ref{scalability_pic}. The trend of the wall-clock time shows that the computational time to complete the training is not affected by the increasing number of data classes, as long as the computational workload for each MPI process stays fixed, thus showing that weak scaling is obtained by the training of distributed DC-CGANs on all three datasets. 

The average GPU usage computed across all the GPUs engaged in the computation is 70\% and the GPU memory occupation is around 31\% for every numerical test performed.

\section{Conclusions and future developments}
We presented a distributed approach to train DC-CGANs models which reduces the imbalance between the computational tasks of the discriminator and the generator. The distribution of the task is based on partitioning the data in classes and it is justified by the variability of features between data classes that prevails over the variability of features within data points of the same class. 
The reduced imbalance results in better images generated by the trained DC-CGANs compared to similar state-of-the-art approaches. 
The results have also shown an almost ideal weak scaling, indicating that HPC computing resources are fully leveraged.  

Our future work aims at extending the study performed in this work to more complex but also more accurate neural network architectures, such as auxiliary classifier GANs (AC-GANs) \cite{pmlr-v70-odena17a}, residual neural networks (ResNet) \cite{Wang2017GenerativeAN}, self-attention generative adversarial neural networks (SAGANs) \cite{pmlr-v97-zhang19d}, and Wasserstein generative adversarial networks (WGANs) \cite{pmlr-v70-arjovsky17a}.

\section*{acknowledgements}
Massimiliano Lupo Pasini thanks Dr. Vladimir Protopopescu for his valuable feedback in the preparation of this manuscript.

Research completed through the Artificial Intelligence Initiative sponsored by the Laboratory Directed Research and Development Program of Oak Ridge National Laboratory, managed by UT-Battelle, LLC, for the U. S. Department of Energy. 

This work used resources of the Oak Ridge Leadership Computing Facility, which is supported by the Office of Science of the U.S. Department of Energy under Contract No. DE-AC05-00OR22725. 

%
\section*{Conflict of interest}
The authors declare that they have no conflict of interest.

\bibliographystyle{unsrt}
\bibliography{references}
\end{document}